\documentclass[15pt a4paper]{article}

\usepackage{latexsym}
\usepackage{amsmath}
\usepackage{amssymb}
\usepackage{graphicx}
\graphicspath{{./}{./figs/}}
\usepackage{algorithm}
\usepackage{algorithmic}
\usepackage{bm}
\usepackage{color}
\usepackage{cite}

\newenvironment{proof}{{\noindent\it Proof.\,}}{\hfill $\square$\par}\newcommand{\trace}{\mathop{\mathrm{tr}}}%
\newcommand{\re}{\mathop{\mathrm{Re}}}%

\begin{document}

\title{\bf Online Learning Algorithms for Quaternion ARMA Model}
\author{Xiaokun Pu, Chunguang Li\thanks{Corresponding author, Email:
cgli@zju.edu.cn}}

\date{\small College of Information Science and Electronic Engineering, \\Zhejiang University, Hangzhou 310027, P. R. China.}

\maketitle

\begin{abstract}
In this paper, we address the problem of adaptive learning for
autoregressive moving average (ARMA) model in the quaternion
domain. By transforming the original learning problem into a full
information optimization task without explicit noise terms, and
then solving the optimization problem using the gradient descent
and the Newton analogues, we obtain two online learning algorithms
for the quaternion ARMA. Furthermore, regret bound analysis
accounting for the specific properties of quaternion algebra is
presented, which proves that the performance of the online
algorithms asymptotically approaches that of the best quaternion
ARMA model in hindsight.
\end{abstract}

\section{Introduction}

{I}{n} recent years, quaternion algebra has attracted considerable attention in the signal processing community. As a natural representation of 3D and 4D signals, quaternion allows for a reduction in the number of parameters and operations involved, and can bring insights that would not be acquired by real- and complex-valued representations. Due to these elegant properties, quaternion adaptive signal processing algorithms have developed rapidly and have achieved satisfactory performance in a wide range of applications
\cite{took2009qlms,took2010wlqlms,took2010qiir,mandic2014qkf,tobar2014,xia2015a,xia2015b,shang2014}.

Despite the existence of many quaternion algorithms, we notice
that so far, there is no learning algorithm for the ARMA model in
the quaternion domain. Due to its flexibility in the modeling of
actual time series, the ARMA has been widely used in the real and
complex domains for modeling 1D and 2D time series
\cite{arma2,arma3,arma4,armaapp2,armaapp3}. Thus, when it comes to
3D and 4D situations, a natural idea is to extend the ARMA model
and its learning algorithms to the quaternion domain in order to
take advantage of quaternion-valued representation.

To this end, in this paper, we propose two online learning
algorithms for the quaternion ARMA (qARMA) model. The online
learning is achieved by borrowing the idea of ``improper
learning'' principle\cite{oarma,oarima} in the real domain, and
transforming the learning problem of qARMA into a full information
optimization task without explicit noise terms. We then solve the
optimization problem by extending the online gradient descent method \cite{ogd} and the online Newton method \cite{ons} to the quaternion domain.
Furthermore, we present
regret bound analysis of the proposed method to illustrate the
validity of this transformation, which, to the best of our
knowledge, is the first time that regret bound analysis for quaternion algorithms is performed. The
theoretical results guarantee that the performance of the online
algorithms asymptotically approaches that of the best quaternion
ARMA model in hindsight.

    \section{Preliminaries} \label{sec:2}

\subsection{Notations} \label{sec:2a}

In this letter, for a quaternion vector $q=q_a+q_bi+q_cj+q_dk \in
\mathbf{H}^{n}$, expressed by its real coordinate vectors $q_a,q_b,q_c$, and $q_d \in \mathbf{R}^{n}$, we use
$q^*$ to denote its conjugate, $q^i,q^j,q^k$ to denote its
involution \cite{involution}, $q^T$ to denote its transpose, and
$q^H$ to denote its Hermintian transpose. We use
$\left\|q\right\|=\sqrt{q^Hq}$ to denote its norm. We use
underlined letter $\underline{q}=[q^T,q^{iT},q^{jT},q^{kT}]^T \in
\mathbf{H}^{4n}$ to denote its augmented quaternion vector
\cite{took2010wlqlms}, and $r_q=[q_a^T,q_b^T,q_c^T,q_d^T]^T
\in \mathbf{R}^{4n}$ to denote its dual-quadrivariate real vector.
The relationship bewteen $\underline{q}$ and $r_q$ is given by \cite{q=jr}
\begin{equation}
\label{h=jr}
\underbrace{\left(
	\begin{array}{cccc}
	q\\q^i\\q^j\\q^k
	\end{array}
	\right)}_{\underline{q}}
=
\underbrace{\left(
	\begin{array}{cccc}
	I_n & iI_n & jI_n & kI_n\\
	I_n & iI_n & -jI_n & -kI_n\\
	I_n & -iI_n & jI_n & -kI_n\\
	I_n & -iI_n & -jI_n & kI_n
	\end{array}
	\right)}_{J}
\underbrace{\left(
	\begin{array}{cccc}
	q_a\\q_b\\q_c\\q_d
	\end{array}
	\right)}_{r_q},
\end{equation}
where $I_n$ is the $n\times n$ identity matrix, and $J$ denotes the $4n\times 4n$ matrix in (\ref{h=jr}). From (\ref{h=jr}), we see that a real scalar function $f(q): \mathbf{H}^n \to \mathbf{R}$ can be viewed in three equivalent forms
\begin{equation}
\nonumber
f(q)\Leftrightarrow f(r_q) \triangleq f(q_a,q_b,q_c,q_d) \Leftrightarrow f(\underline{q}) \triangleq f(q,q^i,q^j,q^k).
\end{equation}

\subsection{Quaternion Gradient} \label{sec:2b}

The usual definition of quaternion derivatives in the mathematical
literature applies only for analytic functions. However, for most
optimization problems, the objective function is real-valued and
thus not analytic.  To this end, the generalized Hamilton-real
(GHR) calculus \cite{ghr} is introduced for defining the
quaternion gradient of nonanalytic functions, given by
\begin{equation}
\nonumber
\nabla_qf\triangleq(\frac{\partial f}{\partial q})^T = (\frac{\partial f}{\partial q_1},...,\frac{\partial f}{\partial q_n})^T \in \mathbf{H}^n.
\end{equation}

According to this definition, for a real scalar function $f(q):
\mathbf{H}^n \to \mathbf{R}$, we have the following equality
relation bewteen the real gradient
$\nabla_{r_q}f\in\mathbf{R}^{4n}$ and the augmented quaternion
gradient $\nabla_{\underline{q}^*}f\in\mathbf{H}^{4n}$
\cite{opInQ}
\begin{equation}
\label{dr=jdq}
\nabla_{r_q}f=J^H\nabla_{\underline{q}^*}f.
\end{equation}

\section{Online Leanring for Quaternion ARMA} \label{sec:3}
In the quaternion domain, a time series is defined as a sequence
of quaternion-valued signals that are observed at successive time
points. Let $x_t \in \mathbf{H}$ denotes the observation at time
$t$, and $\epsilon_t \in \mathbf{H}$ denotes the zero-mean random
noise at time $t$, the qARMA$(p,q)$ model assumes that $x_t$
is generated via the formula
\begin{equation} \nonumber
x_t=\sum_{i=1}^{p} \alpha_i x_{t-i} + \sum_{i=1}^{q} \beta_i
\epsilon_{t-i} + \epsilon_{t},
\end{equation}
where $(p, q)$ is the order of the qARMA model, and $\alpha_i, \beta_i \in \mathbf{H}$ are the quaternion-valued coefficients.
Note that due to the product mechanism of quaternions, each component
of the quaternion signal is correlated to all the four components
in a natural manner.

We now formulate the learning problem of qARMA in a standard online
learning setting. Assume that the data sequence $\{x_t\}^T_{t=1}$ is generated by a qARMA$(p,q)$ model with fixed coefficients. At each time point, we make a prediction
$\tilde{x}_t$, after which the true signal $x_t$ is revealed, and
we suffer a real scalar loss denoted by $l_t(x_t,\tilde{x}_t)$. Then the
\emph{regret} of the prediction is defined as the total prediction loss
minus the total loss of the best possible qARMA model in hindsight
\begin{equation}
\nonumber
R_T=\sum_{t=1}^{T}l_t(x_t,\tilde{x}_t) -
\mathop{\min}_{\alpha,\beta}
\sum_{t=1}^{T}f_t(\alpha,\beta),
\end{equation}
where we define
\begin{equation}
\nonumber
f_t(\alpha,\beta)=l_t(x_t,(\sum_{i=1}^{p}
\alpha_i x_{t-i} + \sum_{i=1}^{q} \beta_i \epsilon_{t-i})).
\end{equation}
Regret measures the difference in performance bewteen the online
prediction and the best fixed model. Then our goal is to design an
efficient online algorithm to minimize this difference, which
guarantees the prediction given by the online algorithm is close to that given by the best qARMA model in
hindsight.

Based on the definition of the regret, an intuitive idea is to
directly estimate the coefficient $(\alpha,\beta)$ of the qARMA
model. Unfortunately, this is difficult due to the existence of
noise terms, which are not revealed to us even in hindsight. To this end,
a possible method is to iteratively estimate the noise terms in the process of coefficient estimation, i.e., innovation-based method. However, it can be foreseen that, similar to the real domain situation \cite{adap}, this method usually requires strong assumptions about the noise terms, such as the most common Guassian assumption.
In this paper, we take an alternative method by borrowing the idea of ``improper learning" principle in the real domain \cite{oarma,oarima} to
transform the learning problem of qARMA into a full information
optimization task without explicit noise terms. Specifically, to
eliminate the unknown noise terms, we approximate the original
qARMA$(p,q)$ model with a qAR$(p+m)$ model, where $m$ is a
properly chosen constant. Then, the loss is given by
\begin{equation}
\begin{split}
\label{l_t^m}
l_t^m(\gamma_t) = l_t(x_t,\tilde{x}_t(\gamma_t)) =l_t(x_t,\sum_{i=1}^{p+m} \gamma_{it} x_{t-i}),
\end{split}
\end{equation}
where $\gamma_{it} \in \mathbf{H}$ is the qAR coefficient at time $t$.

Note that $l_t^m(\gamma_t): \mathbf{H}^{p+m} \to \mathbf{R}$ in
(\ref{l_t^m}) can also be treated as a real mapping
$l_t^m(r_{\gamma_t}): \mathbf{R}^{4(p+m)} \to \mathbf{R}$. Thus,
classical optimization methods can be used to learn the model. In
the sequel, we focus on two popular online optimization
algorithms, Online Gradient Descent (ODG) \cite{ogd} and Online
Newton Step (ONS) \cite{ons}. We extend these algorithms to the
quaternion domain to learn the qAR approximation of the qARMA
model. In the next section, we give regret bound analysis to
demonstrate the validity of this approximation.

\subsection{Quaternion Online Gradient Descent for qARMA}
Online Gradient Descent is a first-order optimization algorithm in
the real domain, which finds the optimal point by taking a step
proportional to the negative of the gradient of the instantaneous
loss function at each iteration. For the learning problem of
qARMA, since the instantaneous loss $l_t^m(\gamma_t):
\mathbf{H}^{p+m} \to \mathbf{R}$ can be treated as a real mapping
$l_t^m(r_{\gamma_t}): \mathbf{R}^{4(p+m)} \to \mathbf{R}$, then
according to OGD, we have the following quadrivariate real
gradient descent update rule \cite{ogd}
\begin{equation}
\label{deltar}
\Delta r_{\gamma_{t}} = \eta(-\nabla_{r_{\gamma_t}}l_t^m),
\end{equation}
where $\Delta$ denotes a small increment and $\eta \in
\mathbf{R}^+$ is the learning rate. In the sequel, we omit $l_t^m$
in the quaternion gradient for simplicity. According to
(\ref{h=jr}), we have the increment expression for the augmented
vector \underline{$\gamma$} given by
\begin{equation}
\label{delta h=jr}
\Delta \underline{\gamma_{t}} = J\Delta r_{\gamma_{t}}.
\end{equation}
Plugging  (\ref{deltar}) into (\ref{delta h=jr}) yields
\begin{equation}
\nonumber
\Delta \underline{\gamma_{t}} = -\eta J \nabla_{r_{\gamma_t}}.
\end{equation}
Next according to the equality relation (\ref{dr=jdq}), we have
\begin{equation}
\nonumber
\Delta \underline{\gamma_{t}} = -\eta J J^H \nabla_{\underline{\gamma_t^*}}.
\end{equation}
Applying the fact that $JJ^H=4I_{4(p+m)}$ and the correspondence between quaternion vector and its augmented vector, we get the quaternion online gradient descent update rule in the form
\begin{equation}
\label{qogd}
\Delta {\gamma_{t}}  = -4\eta \nabla_{{\gamma_t^*}}.
\end{equation}

Based on (\ref{qogd}), we now obtain the Quaternion Online
Gradient Descent algorithm for qARMA summarized in Algorithm 1,
where $\mathcal{{K}}$ refers to the decision set of the
coefficient vector $\gamma$, i.e., $\mathcal{{K}}=\{{\gamma} \in
\mathbf{H}^{p+m}, \left| \gamma_j \right| \leq c,j=1,...,p+m\}$,
and $\prod_{\mathcal{{K}}}({y})$ refers to the Euclidean
projection onto $\mathcal{{K}}$, i.e.,
$\prod_{\mathcal{{K}}}({y})=\arg\min_{{x} \in \mathcal{{K}}}
\left\|{y}-{x}\right\|_2$. The projection step here ensures that
$\gamma_{t}$ is always in the feasible region.
\begin{algorithm}[htb]
	\caption{qARMA-QOGD(p,q)}
	\textbf{{Initialization:}}
	qARMA order $(p,q)$; learning rate $\eta \in \mathbf{R}^+$.
	
	set $m \ge \log_{\lambda_{max}} ((TLM_{max}q)^{-1})$.
	
	choose $\gamma_1 \in \mathcal{K}$ arbitrarily.
	
	\textbf{for} $t=1$ \textbf{to} $T$ \textbf{do}
	
	\qquad prediction: $\tilde{x}_t(\gamma_t) = \sum_{i=1}^{p+m} \gamma_{it} x_{t-i}$;
	
	\qquad loss calculation: $l_t^m(\gamma_t) = l_t(x_t,\sum_{i=1}^{p+m} \gamma_{it} x_{t-i})$;
	
	\qquad gradient calculation: $\nabla_{\gamma_t^*} = \partial l_t^m(\gamma_t)/\partial{\gamma_t^*}$;
	
	\qquad descent: $\phi_{t+1}=\gamma_{t}-4\eta \nabla_{{\gamma_t^*}}$;
	
	\qquad projection: ${\gamma_{t+1}} \gets \prod_{\mathcal{{K}}} (\phi_{t+1})$;
	
	\textbf{end for}
	
\end{algorithm}

\emph{Remark 1:} Note that the online optimization problem can be
solved directly using the quadrivariate real gradient descent
update rule in (\ref{deltar}). However, as discussed in
\cite{opInQ}, since the original problem is quaternion valued, it
is often awkward to reformulate the problem in the real domain and
very tedious to calculate gradients for the optimization in even
moderately complex quaternion dynamic systems. On the contrary,
the quaternion online gradient descent update rule in the
quaternion domain is elegant and easy to  calculate using the GHR
calculus \cite{ghr}.

\subsection{Quaternion Online Newton Step for qARMA}
Online Newton Step is a second-order online optimization algorithm, which uses an approximation of Hessian to obtain better descent directions than first-order optimization algorithms. Similar to OGD, for the learning problem of qARMA, we have the following Newton iteration step for quadrivariate real vector according to ONS \cite{ons}
\begin{equation}
\label{drons}
\Delta r_{\gamma_{t}} = \eta(- A_t^{-1} \nabla_{r_{\gamma_t}}),
\end{equation}
where the matrix $A_t=\sum_{i=1}^{t}\nabla_{r_{\gamma_i}}\nabla_{r_{\gamma_i}}^H$ is related to the real Hessian as discussed in \cite{ons}.
Plugging (\ref{drons}) into (\ref{delta h=jr}) yields
\begin{equation}
\nonumber
\begin{split}
\Delta \underline{\gamma_{t}}  &= -\eta J A_t^{-1} \nabla_{r_{\gamma_t}} = -\eta J (\sum_{i=1}^{t}\nabla_{r_{\gamma_i}}\nabla_{r_{\gamma_i}}^H)^{-1} \nabla_{r_{\gamma_t}}.
\end{split}
\end{equation}
Then according to (\ref{dr=jdq}), we have
\begin{equation}
\begin{split}
\nonumber
\Delta \underline{\gamma_{t}}
= -\eta J (\sum_{i=1}^{t}J^H \nabla_{\underline{\gamma_t^*}} \nabla_{\underline{\gamma_t^*}}^H J)^{-1} J^H \nabla_{\underline{\gamma_t^*}}.
\end{split}
\end{equation}
Using the fact that $JJ^H=4I_{4(p+m)}$, and that $(AB)^{-1}=B^{-1}A^{-1}$ generally holds for invertible quaternion matrix $A$ and $B$, we get the quaternion online Newton step update rule in the form
\begin{equation}
\begin{split}
\label{qons}
\Delta \underline{\gamma_{t}}
=-\eta  (\sum_{i=1}^{t} \nabla_{\underline{\gamma_t^*}} \nabla_{\underline{\gamma_t^*}}^H )^{-1}  \nabla_{\underline{\gamma_t^*}}
=-\eta  A_{qt}^{-1}  \nabla_{\underline{\gamma_t^*}},
\end{split}
\end{equation}
where we denote $A_{qt}\triangleq\sum_{i=1}^{t} \nabla_{\underline{\gamma_i^*}} \nabla_{\underline{\gamma_i^*}}^H$.

Based on (\ref{qons}), we now obtain the Quaternion Online Newton
Step algorithm for qARMA summarized in Algorithm 2, where
$\mathcal{\underline{K}}$ refers to the corresponding decision set
of the augmented vector $\underline{\gamma}$,
$\prod_{\mathcal{\underline{K}}}^{A_{qt}}(\underline{y})$ refers
to the Euclidean projection onto $\mathcal{\underline{K}}$ in the
norm induced by $A_{qt}$, i.e.,
$\prod_{\mathcal{\underline{K}}}^{A_{qt}}(\underline{y})=\arg\min_{\underline{x}
	\in \mathcal{\underline{K}}}
(\underline{y}-\underline{x})^HA_{qt}(\underline{y}-\underline{x})$,
and $P$ is the $\mathbf{R}^{n\times 4n}$ matrix in (\ref{T}) which
gives the relation bewteen $\gamma_t$ and its augmented vector
\underline{$\gamma_t$} by
\begin{equation}
\label{T}
\gamma_t = \underbrace{\left(
	\begin{array}{cccc}
	I_n & 0_{n\times 3n}
	\end{array}
	\right)}_{P}\underline{\gamma_t}.
\end{equation}

\begin{algorithm}[htb]
	\caption{qARMA-ONS(p,q)}
	\textbf{{Initialization:}}
	qARMA order $(p,q)$; learning rate $\eta \in \mathbf{R}^+$; initial matrix $A_{q0} \in \mathbf{H}^{4(p+m)\times 4(p+m)}$.
	
	set $m \ge \log_{\lambda_{max}} ((TLM_{max}q)^{-1})$.
	
	choose $\gamma_1 \in \mathcal{K}$ arbitrarily.
	
	\textbf{for} $t=1$ \textbf{to} $T$ \textbf{do}
	
	\qquad prediction: $\tilde{x}_t(\gamma_t) = \sum_{i=1}^{p+m} \gamma_{it} x_{t-i}$;
	
	\qquad loss calculation: $l_t^m(\gamma_t) = l_t(x_t,\sum_{i=1}^{p+m} \gamma_{it} x_{t-i})$;
	
	\qquad gradient calculation: $\nabla_{\gamma_t^*} = \partial l_t^m(\gamma_t)/\partial{\gamma_t^*}$;
	
	\qquad matrix update: $A_{qt} \gets A_{q(t-1)} + \nabla_{\underline{\gamma_t^*}} \nabla_{\underline{\gamma_t^*}}^H$;
	
	\qquad descent: $\underline{\phi_{t+1}} =  \underline{\gamma_t}-\eta A_{qt}^{-1} \nabla_{\underline{\gamma_t^*}}$;
	
	\qquad projection: ${\gamma_{t+1}} \gets P\prod_{\mathcal{\underline{K}}}^{A_{qt}} (\underline{\phi_{t+1}})$;
	
	\textbf{end for}
\end{algorithm}

\emph{Remark 2:} Note that $A_{qt}^{-1}$ can be calculated incrementally using the Sherman-Morrison formula, thus Algorithm 2 can be performed efficiently in an online manner.

\section{Regret Bound Analysis} \label{sec:4}
In this section, we present regret bound analyses for the proposed
algorithms. Some necessary assumptions are listed below. We remark
that here the Guassian assumption about the noise terms is not
required, which means that the algorithms are applicable to the
scenarios with non-Guassian noise terms.
\begin{enumerate}
	\item[1.] The coefficient $\alpha$ satisfies $\left|\alpha_i\right| \leq c$ for some $c \in \mathbf{R}^+$.
	\item[2.] The coefficient $\beta$ satisfies that a $q$-th order difference equation with coefficients $\left|\beta_1\right|,\left|\beta_2\right|,...,\left|\beta_q\right|$ and real-valued observations is a stationary process.
	\item[3.] The noises are stochastically and independently generated. Also we assume $\mbox{E}\left[ \left|\epsilon_t\right| \right] < M_{max} < \infty$ and $\mbox{E}\left[ l_t(x_t,x_t-\epsilon_t) \right] < \infty$.
	\item[4.] The loss function $l_t$ is Lipschitz continuous with Lipschitz constant $L\in\mathbf{R}^+$.
	\item[5.] The loss functions in (\ref{l_t^m}) have bounded augmented decision set $\underline{\mathcal{K}}$, i.e., $\forall \underline{\gamma_1},\underline{\gamma_2}\in \underline{\mathcal{K}}$, $\|\underline{\gamma_1}-\underline{\gamma_2}\|\leq D$, and have bounded augmented quaternion gradient, i.e., $\forall \underline{\gamma}\in
	\underline{\mathcal{K}}$, $\|\nabla_{\underline{\gamma}}{l_t^m}\| \leq G$.
\end{enumerate}

For the analysis of Algorithm 1, we also assume that
\begin{enumerate}
	\item[6.] The loss functions in (\ref{l_t^m}) are $H$-strong convex for some $H\in\mathbf{R}^+$,
	i.e.,
	\begin{equation}
	\label{H-convex}
	\forall \gamma \in \mathcal{K}, \, H_{\underline{\gamma}\underline{\gamma^*}}(l_t^m) \succeq HI_{4(p+m)},
	\end{equation}
	where $H_{\underline{\gamma}\underline{\gamma^*}}(l_t^m)$ is the  augmented quaternion Hessian matrix introduced in \cite{opInQ}.
\end{enumerate}

For the analysis of Algorithm 2, we can relax the $H$-strong convex assumption to $\lambda$-exp-concave, that is
\begin{enumerate}
	\item[7.] The loss functions in (\ref{l_t^m}) are  $\lambda$-exp-concave for some $\lambda\in\mathbf{R}^+$, i.e.,
	\begin{equation}
	\label{exp-convex}
	\forall \gamma \in \mathcal{K}, \, H_{\underline{\gamma}\underline{\gamma^*}}[\exp(-\lambda l_t^m)] \preceq 0_{4(p+m)}.
	\end{equation}
\end{enumerate}

\subsection{Validity of qAR Approximation} \label{sec:4a}

As discussed in Section 3, the main difficulty of qARMA
learning is the existence of the unknown noise terms. To this end,
since a qARMA$(p,q)$ process is equivalent to a qAR$(\infty)$
process, we recursively define
\begin{equation} \nonumber
x_t^{\infty}(\alpha,\beta)=\sum_{i=1}^{p}\alpha_i x_{t-i} + \sum_{i=1}^{q}\beta_i(x_{t-i}-x_{t-i}^{\infty}(\alpha,\beta))
\end{equation}
with initial condition $x_1^{\infty}(\alpha,\beta)=x_1$ by using the entire past historical data to eliminate the explicit noise terms.
Then for practical consideration, we truncate the memory length and recursively define
\begin{equation} \nonumber
x_t^{m}(\alpha,\beta)=\sum_{i=1}^{p}\alpha_i x_{t-i} + \sum_{i=1}^{q}\beta_i(x_{t-i}-x_{t-i}^{m-i}(\alpha,\beta))
\end{equation}
with initial condition $x_t^{m}(\alpha,\beta)=x_t$ for all $t$ and
$m \leq 0$ to use only the last $p+m$ data points. Note that in
substance, $x_t^m$ is a qAR$(p+m)$ process, i.e., we naturally
transform the learning problem of qARMA into a finite-order qAR
learning problem through the above two definitions. Then the rest
is to vertify the validity of this transformation.

To this end, we adopt the difference equation technique introduced
in \cite{oarima} for the analysis of real ARMA model. We
demonstrate that this technique is also applicable to the
quaternion situation.  For simplicity, we define
\begin{equation}
\nonumber
\begin{split}
&f_t^{\infty}(\alpha,\beta) = l_t(x_t,x_t^{\infty}(\alpha,\beta)),\\
&f_t^{m}(\alpha,\beta) = l_t(x_t,x_t^{m}(\alpha,\beta)),
\end{split}
\end{equation}
and let $(\alpha^\star,\beta^\star) = \arg\min_{\alpha,\beta}
\sum_{t=1}^{T}\mbox{E}\left[f_t(\alpha,\beta)\right]$ denote the
best qARMA coefficient in hindsight. We then have the following
Lemma 2-4 about the relation bewteen the loss of the qAR
prediction and that of the qARMA prediction.

{\emph{Lemma 1}}\cite{oarima}: Given Assumption 2 that a $q$-th order difference equation with coefficients $|\beta_1|,...,|\beta_q|$ and \emph{real-valued} observations $\{y_t\in \mathbf{R}\}_{t=-(q-1)}^{T}$ is a stationary process, $\lambda_1,...,\lambda_q$ are the q roots of this AR characteristic equation. Let we set $\lambda_{max}=\max\{|\lambda_1|,...,|\lambda_q|\}$, it holds that
\begin{equation}
y_t \leq \lambda_{max}^t(y_0+y_{-1}+...+y_{-(q-1)}). \nonumber
\end{equation}

{\emph{Lemma 2}}: For the quaternion data sequence
$\{x_t\}_{t=1}^T$ generated by any qARMA model satisfying
Assumption 1-3 and the loss function satisfying Assumption 4, it
holds that
\begin{equation} \nonumber
\min_{\gamma}\sum_{t=1}^{T}l_t^m(\gamma) \leq \sum_{t=1}^{T}f_t^m(\alpha^\star,\beta^\star).
\end{equation}

\begin{proof}
	Note that if we let $\gamma^\star=c(\alpha^\star,\beta^\star)$ be the corresponding qAR coefficient of the $x_t^m(\alpha^\star,\beta^\star)$ process, we immediately get that
	\begin{equation} \nonumber
	\sum_{t=1}^{T}l_t^m(\gamma^\star)=\sum_{t=1}^{T}f_t^m(\alpha^\star,\beta^\star).
	\end{equation}
	Trivially, it always holds that
	\begin{equation} \nonumber
	\min_{\gamma}\sum_{t=1}^{T}l_t^m(\gamma) \leq
	\sum_{t=1}^{T}l_t^m(\gamma^\star).
	\end{equation}
	Combining the above two equations, we complete the proof.
\end{proof}

{\emph{Lemma 3}}: For the quaternion data sequence
$\{x_t\}_{t=1}^T$ generated by any qARMA model satisfying
Assumption 1-3 and the loss function satisfying Assumption 4, it
holds that
\begin{equation} \nonumber
\left|\sum_{t=1}^{T}\mbox{E}\left[f_t^{\infty}(\alpha^\star,\beta^\star)\right]-
\sum_{t=1}^{T}\mbox{E}\left[f_t(\alpha^\star,\beta^\star)\right]\right|
=O(1).
\end{equation}

\begin{proof}
	We begin the proof by analyzing $\mbox{E}[|x_t-x_t^{\infty}(\alpha^\star,\beta^\star)-\epsilon_t|]$.
	\begin{equation} \nonumber
	\begin{split}
	&\mbox{E}[|x_t-x_t^{\infty}(\alpha^\star,\beta^\star)-\epsilon_t|]\\
	&= \mbox{E}[|\sum_{i=1}^{p}\alpha_i^\star x_{t-i}+\sum_{i=1}^{q}\beta_i^\star\epsilon_{t-i}   +\epsilon_t - \sum_{i=1}^{p}\alpha_i^\star x_{t-i} - \sum_{i=1}^{q}\beta_i^\star(x_{t-i}-x_{t-i}^{\infty}(\alpha^\star,\beta^\star)) -\epsilon_t|] \\
	&= \mbox{E}[| \sum_{i=1}^{q}\beta^\star(x_{t-i}^{\infty}(\alpha^\star,\beta^\star)-x_{t-i}+\epsilon_{t-i}) |] \leq \sum_{i=1}^{q} |\beta_i^\star|  \mbox{E}[|x_{t-i}-x_{t-i}^{\infty}(\alpha^\star,\beta^\star)-\epsilon_{t-i} |].
	\end{split}
	\end{equation}
	Based on the above inequality, Assumption 2, and Lemma 1, we have
	\begin{equation} \nonumber
	\begin{split}
	&\mbox{E}[|x_t-x_t^{\infty}(\alpha^\star,\beta^\star)-\epsilon_t|]\\
	&\leq \lambda_{max}^t( \mbox{E}[|x_0-x_0^{\infty}(\alpha^\star,\beta^\star)-\epsilon_{0}|]  +...+ \mbox{E}[|x_{-(q-1)}-x_{-(q-1)}^{\infty}(\alpha^\star,\beta^\star)-\epsilon_{-(q-1)}|])=\lambda_{max}^t\rho,
	\end{split}
	\end{equation}
	where we use $\rho$ to represent the summation in the above
	bracket for simplicity. According to Lemma 1, we know
	$|\lambda_{max}|<1$ for a stationary process, which means that
	$\mbox{E}[|x_t-x_t^{\infty}(\alpha^\star,\beta^\star)-\epsilon_t|]$
	decays exponentially as $t$ increasing linearly.
	
	From Assumptions 3, we know that $\epsilon_t$ is stochastic and
	independent of $\epsilon_1,...,\epsilon_{t-1}$ and hence the best
	prediction available at time $t$ will cause a loss no less than
	$l_t(x_t,x_t-\epsilon_t)$. Recall that $l_t$ is assumed to be
	Lipschitz continuous with Lipschitz constant  $L\in\mathbf{R}^+$
	in Assumption 4, we have that
	\begin{equation} \nonumber
	\begin{split}
	&|\mbox{E}\left[f_t^{\infty}(\alpha^\star,\beta^\star)\right]-
	\mbox{E}\left[f_t(\alpha^\star,\beta^\star)\right]| \\
	&=|\mbox{E}[l_t(x_t,x_t^{\infty}(\alpha^\star,\beta^\star))]-
	\mbox{E}[l_t(x_t,x_t-\epsilon_t)]| \\&
	\leq \mbox{E}[|l_t(x_t,x_t^{\infty}(\alpha^\star,\beta^\star))-
	l_t(x_t,x_t-\epsilon_t)|]\\
	& \leq L\cdot \mbox{E}[|x_t-x_t^{\infty}(\alpha^\star,\beta^\star)-\epsilon_t|] \leq L\cdot \lambda_{max}^t\rho,
	\end{split}
	\end{equation}
	where the first inequlity follows from Jensen's inequality. By summing the above for all $t$ we get that
	\begin{equation} \nonumber
	\left|\sum_{t=1}^{T}\mbox{E}\left[f_t^{\infty}(\alpha^\star,\beta^\star)\right]-
	\sum_{t=1}^{T}\mbox{E}\left[f_t(\alpha^\star,\beta^\star)\right]\right|
	=O(1).
	\end{equation}
\end{proof}

{\emph{Lemma 4}}: For the quaternion data sequence
$\{x_t\}_{t=1}^T$ generated by any qARMA model satisfying
Assumption 1-3 and the loss function satisfying Assumption 4, if
we choose $m \ge \log_{\lambda_{max}} ((TLM_{max}q)^{-1})$, then
we have
\begin{equation} \nonumber
\left|\sum_{t=1}^{T}\mbox{E}\left[f_t^m(\alpha^\star,\beta^\star)\right]-
\sum_{t=1}^{T}\mbox{E}\left[f_t^{\infty}(\alpha^\star,\beta^\star)\right]\right|
=O(1).
\end{equation}
\begin{proof}
	For arbitrary $t$, we focus on the distance between
	$f_t^{\infty}(\alpha^\star,\beta^\star)$ and
	$f_t^m(\alpha^\star,\beta^\star)$ in expectation. First, for any
	$m \in \{0,-1,...,-(1-q)\}$ we have
	$x_t^m(\alpha^\star,\beta^\star)=x_t$ by definition, and hence
	\begin{equation} \nonumber
	\begin{split}
	&\left|x_t^m(\alpha^\star,\beta^\star)-x_t^{\infty}(\alpha^\star,\beta^\star)\right| = \left|x_t-x_t^{\infty}(\alpha^\star,\beta^\star)\right| \leq \left|x_t-x_t^{\infty}(\alpha^\star,\beta^\star)-\epsilon_t\right| +\left|\epsilon_t\right|.
	\end{split}
	\end{equation}
	From Assumption 3, we know that $\mbox{E}\left[
	\left|\epsilon_t\right| \right] < M_{max} < \infty$ for all $t$,
	and we know that
	$\mbox{E}[|x_t-x_t^{\infty}(\alpha^\star,\beta^\star)-\epsilon_t|]$
	decays exponentially as proven in Lemma 3, and hence we have
	$\mbox{E}[\left|x_t^m(\alpha^\star,\beta^\star)-x_t^{\infty}(\alpha^\star,\beta^\star)\right|]<2M_{max}$.
	Next, we show that
	$\left|x_t^m(\alpha^\star,\beta^\star)-x_t^{\infty}(\alpha^\star,\beta^\star)\right|$
	exponentially decreases as $m$ increases linearly.
	\begin{equation} \nonumber
	\begin{split}
	&\left|x_t^m(\alpha^\star,\beta^\star)-x_t^{\infty}(\alpha^\star,\beta^\star)\right| \\
	&=| \sum_{i=1}^{q}\beta_i^\star(x_{t-i}-x_{t-i}^{m-i}(\alpha^\star,\beta^\star)) - \sum_{i=1}^{q}\beta_i^\star(x_{t-i}-x_{t-i}^{\infty}(\alpha^\star,\beta^\star))   |   \\
	&= |\sum_{i=1}^{q}\beta_i^\star(x_{t-i}^{m-i}(\alpha^\star,\beta^\star)-x_{t-i}^{\infty}(\alpha^\star,\beta^\star))| \\&
	\leq \sum_{i=1}^{q}|\beta_i^\star||x_{t-i}^{m-i}(\alpha^\star,\beta^\star)-x_{t-i}^{\infty}(\alpha^\star,\beta^\star)|.
	\end{split}
	\end{equation}
	Based on the above inequality, Assumption 2, and Lemma 1, we have
	\begin{equation} \nonumber
	\begin{split}
	&\left|x_t^m(\alpha^\star,\beta^\star)-x_t^{\infty}(\alpha^\star,\beta^\star)\right| \\
	& \leq \lambda_{max}^m(
	|x_{t-m}^0(\alpha^\star,\beta^\star)-x_{t-m}^{\infty}(\alpha^\star,\beta^\star)|+ ...+ |x_{t-m-(q-1)}^{-(q-1)}(\alpha^\star,\beta^\star)-x_{t-m-(q-1)}^{\infty}(\alpha^\star,\beta^\star)|
	) \leq 2qM_{max}\lambda_{max}^m.
	\end{split}
	\end{equation}
	Recall that $l_t$ is assumed to be Lipschitz continuous with
	Lipschitz constant $L\in\mathbf{R}^+$ in Assumption 4, we have
	\begin{equation} \nonumber
	\begin{split}
	&|\mbox{E}\left[f_t^m(\alpha^\star,\beta^\star)\right]-
	\mbox{E}\left[f_t^{\infty}(\alpha^\star,\beta^\star)\right]| \\
	&=|\mbox{E}[l_t(x_t,x_t^m(\alpha^\star,\beta^\star))]-
	\mbox{E}[l_t(x_t,x_t^{\infty}(\alpha^\star,\beta^\star))]| \\&
	\leq \mbox{E}[|l_t(x_t,x_t^m(\alpha^\star,\beta^\star))-
	l_t(x_t,x_t^{\infty}(\alpha^\star,\beta^\star))|]\\
	& \leq L\cdot \mbox{E}[|x_t^m(\alpha^\star,\beta^\star)-x_t^{\infty}(\alpha^\star,\beta^\star)|] \leq L\cdot 2qM_{max}\lambda_{max}^m.
	\end{split}
	\end{equation}
	Summing the above for all $t$
	results in
	\begin{equation} \nonumber
	\begin{split}
	&\left|\sum_{t=1}^{T}\mbox{E}\left[f_t^m(\alpha^\star,\beta^\star)\right]-
	\sum_{t=1}^{T}\mbox{E}\left[f_t^{\infty}(\alpha^\star,\beta^\star)\right]\right|\leq TL\cdot 2qM_{max}\lambda_{max}^m.
	\end{split}
	\end{equation}
	Finally, by choosing  $m \ge \log_{\lambda_{max}}
	((TLM_{max}q)^{-1})$, we get
	\begin{equation} \nonumber
	\left|\sum_{t=1}^{T}\mbox{E}\left[f_t^m(\alpha^\star,\beta^\star)\right]-
	\sum_{t=1}^{T}\mbox{E}\left[f_t^{\infty}(\alpha^\star,\beta^\star)\right]\right|
	=O(1).
	\end{equation}
\end{proof}

Based on  Lemma 2-4, we have the following Lemma 5 which
guarantees that the performance of the best qAR$(p+m)$ model in hindsight
is close to that of the best qARMA$(p,q)$ model in hindsight on the
average, for some properly chosen constant $m$.

{\emph{Lemma 5}}: For the quaternion data sequence
$\{x_t\}_{t=1}^T$ generated by any qARMA model satisfying
Assumption 1-3 and the loss function satisfying Assumption 4, if
we choose $m \ge \log_{\lambda_{max}} ((TLM_{max}q)^{-1})$, then
we have
\begin{equation}
\nonumber
\min_{\gamma}\sum_{t=1}^{T}l_t^m(\gamma)-
\sum_{t=1}^{T}\mbox{E}\left[f_t(\alpha^\star,\beta^\star)\right]
=O(1).
\end{equation}

\begin{proof}
	According to Lemma 3 and Lemma 4, we have
	\begin{equation*} 
	\sum_{t=1}^{T}\mbox{E}\left[f_t^{\infty}(\alpha^\star,\beta^\star)\right]-
	\sum_{t=1}^{T}\mbox{E}\left[f_t(\alpha^\star,\beta^\star)\right]
	=C_1,
	\end{equation*}
	\begin{equation} \nonumber
	\sum_{t=1}^{T}\mbox{E}\left[f_t^m(\alpha^\star,\beta^\star)\right]-
	\sum_{t=1}^{T}\mbox{E}\left[f_t^{\infty}(\alpha^\star,\beta^\star)\right]
	=C_2,
	\end{equation}
	where $C_1,C_2\in\mathbf{R}$ are constants. Combining the above
	two equations yields
	\begin{equation} \nonumber
	\sum_{t=1}^{T}\mbox{E}\left[f_t^m(\alpha^\star,\beta^\star)\right]-
	\sum_{t=1}^{T}\mbox{E}\left[f_t(\alpha^\star,\beta^\star)\right]
	=C,
	\end{equation}
	where $C=C_1+C_2$ is a constant. From Lemma 2 we know that
	\begin{equation} \nonumber
	\min_{\gamma}\sum_{t=1}^{T}l_t^m(\gamma) \leq \sum_{t=1}^{T}f_t^m(\alpha^\star,\beta^\star)= \sum_{t=1}^{T}\mbox{E}\left[f_t^m(\alpha^\star,\beta^\star)\right].
	\end{equation}
	Combining the above two equations yields
	\begin{equation} \nonumber
	\min_{\gamma}\sum_{t=1}^{T}l_t^m(\gamma) - \sum_{t=1}^{T}\mbox{E}\left[f_t(\alpha^\star,\beta^\star)\right] \leq C.
	\end{equation}
	Thus, we complete the proof of Lemma 5.
\end{proof}

\subsection{Regret Bound Analysis for qARMA-QOGD} \label{sec:4b}
In this subsection, we perform the regret bound analysis for
Algorithm 1, borrowing the idea of \cite{oarma} in the real
domain.

In the following Lemma 6, we bound the regret between the online
qAR$(p+m)$ model and the best qAR$(p+m)$ model in hindsight.

\emph{Lemma 6:} For the quaternion data sequence
$\{x_t\}_{t=1}^T$ generated by any qARMA model satisfying
Assumption 1-3, Algorithm 1 with loss function satisfying
Assumption 4-6 and learning rate satisfying
$\eta_t=\frac{1}{H\cdot t}$ can generate an online quaternion
sequence $\{\gamma_t\}_{t=1}^T$ such that
\begin{equation}
\nonumber
\sum_{t=1}^{T}l_t^m(\gamma_t)-\min_{\gamma} \sum_{t=1}^{T}l_t^m(\gamma)  = O(\frac{G^2}{H}\log T).
\end{equation}

\begin{proof}
	Let $\gamma^\star=\arg\min_{\gamma} \sum_{t=1}^{T}l_t^m(\gamma)$
	denote the best qAR coefficient in hindsight, and denote
	$\nabla_{\gamma} = \nabla_{\gamma} l_t^m$. By using the quaternion
	Taylor series expansion introduced in \cite{opInQ}, we have
	\begin{equation}
	\begin{split}
	\nonumber
	l_t^m(\gamma^\star)&=l_t^m(\gamma_t)+\nabla_{\underline{\gamma_t^*}}^H (\underline{\gamma^\star}-\underline{\gamma_t}) + \frac{1}{2}(\underline{\gamma^\star}-\underline{\gamma_t})^H H_{\underline{\gamma_t\gamma_t^*}}(\underline{\gamma^\star}-\underline{\gamma_t})\\
	&\ge l_t^m(\gamma_t)+\nabla_{\underline{\gamma_t^*}}^H (\underline{\gamma^\star}-\underline{\gamma_t}) + \frac{H}{2}\|\underline{\gamma^\star}-\underline{\gamma_t}\|^2,
	\end{split}
	\end{equation}
	where $\nabla_{\underline{\gamma_t^*}}$ is the augmented quaternion gradient and $H_{\underline{\gamma_t\gamma_t^*}}$ is the augmented Hessian defined in \cite{opInQ}. The inequality above follows from $H$-strong convexity in (\ref{H-convex}). It then follows that
	\begin{gather}
	\label{12}
	l_t^m(\gamma_t)-l_t^m(\gamma^\star)\leq \nabla_{\underline{\gamma_t^*}}^H (\underline{\gamma_t}-\underline{\gamma^\star})-\frac{H}{2}\|\underline{\gamma^\star}-\underline{\gamma_t}\|^2.
	\end{gather}
	Next, according to the descent step and the projection step, we have
	\begin{equation}
	\begin{split}
	\nonumber
	\| \underline{\gamma_{t+1}}-\underline{\gamma^\star} \|^2 &\leq \| \underline{\phi_{t+1}}-\underline{\gamma^\star} \|^2 = \| \underline{\gamma_{t}}-\underline{\gamma^\star}-\eta_{t}\nabla_{\underline{\gamma_t^*}} \|^2,
	\end{split}
	\end{equation}
	and hence
	\begin{equation}
	\begin{split}
	\nonumber
	&\| \underline{\gamma_{t+1}}-\underline{\gamma^\star} \|^2 \leq  \| {\underline{\gamma_{t}}}-\underline{\gamma^\star} \|^2 + {\eta_{t}^2} \| \nabla_{\underline{\gamma_t^*}} \|^2 - 2{\eta_{t}} \nabla_{\underline{\gamma_t^*}}^H (\underline{\gamma_t}-\underline{\gamma^\star}),\\
	& \nabla_{\underline{\gamma_t^*}}^H (\underline{\gamma_t}-\underline{\gamma^\star}) \leq  \frac{1}{2\eta_t} (\| {\underline{\gamma_{t}}}-\underline{\gamma^\star} \|^2 - \| \underline{\gamma_{t+1}}-\underline{\gamma^\star} \|^2 )+\frac{\eta_tG^2}{2}.
	\end{split}
	\end{equation}
	
	According to the above equation, by summing both sides of (\ref{12}) for all $t$ and setting $\eta_t=\frac{1}{H\cdot t}$, we get that
	\begin{equation}
	\begin{split}
	\nonumber
	&\sum_{t=1}^{T}(l_t^m(\gamma_t)-l_t^m(\gamma^\star)) \leq \frac{H}{2}\sum_{t=1}^{T}((t-1)\|{\underline{\gamma_{t}}-\underline{\gamma^*}}\|^2-t\|{\underline{\gamma_{t+1}}-\underline{\gamma^*}}\|^2)+\sum_{t=1}^{T}\frac{G^2}{2Ht}
	\leq \frac{G^2}{2H}(1+\log T).\\
	\end{split}
	\end{equation}
	Thus, we complete the proof of Lemma 6.
\end{proof}

Combining the results in Lemma 5 \& 6, we have the following
Theorem 1, which states a logarithmic bound on the regret of
Algorithm 1. This sublinear regret guarantees that the performance of Algorithm 1 asymptotically approaches that of the best quaternion
ARMA model in hindsight.

{\emph{Theorem 1}}: For the quaternion data sequence
$\{x_t\}_{t=1}^T$ generated by any qARMA model satisfying
Assumption 1-3, Algorithm 1 with loss function satisfying
Assumption 4-6 and learning rate satisfying
$\eta_t=\frac{1}{H\cdot t}$ can generate an online quaternion
sequence $\{\gamma_t\}_{t=1}^T$ such that
\begin{equation}
\nonumber
\label{rb_ogd}
\sum_{t=1}^{T}l_t^m(\gamma_t)-\mathop{\min}_{\alpha,\beta} \sum_{t=1}^{T} \mbox{E}\left[f_t(\alpha,\beta)\right] = O(\frac{G^2}{H}\log T).
\end{equation}

\subsection{Regret Bound of QARMA-QONS} \label{sec:4c}
Regret bound analysis for Algorithm 2 is similar to that of
Algorithm 1, but  is more tedious due to the use of second-order
information. In order to get a theoretical result similar to
Lemma 6, we first introduce several lemmas.

\emph{Lemma 7}: Assume that for $\mathcal{K}\subseteq\mathbf{H}^n$
whose augmented set $\mathcal{\underline{K}}$ has a diameter $D$,
and for all $t$, function $f(q): \mathcal{K}\to\mathbf{R}$ is
$\lambda$-exp-concave and has the property that  $\forall
q\in\mathcal{K},\|\nabla_{\underline{q}}f\|\leq G$. Then for
$\frac{1}{\eta} \leq \frac{1}{2}\min\{\frac{1}{4GD},\lambda\}$, it
holds that
\begin{equation}
\nonumber
\begin{split}
\forall q_1,q_2 \in \mathcal{K},\quad f(q_1) &\ge
f(q_2)+\nabla_{\underline{q_2^*}}^H (\underline{q_1}-\underline{q_2})+\frac{1}{2\eta}(\underline{q_1}-\underline{q_2})^H\nabla_{\underline{q_2^*}} \nabla_{\underline{q_2^*}}^H(\underline{q_1}-\underline{q_2}),
\end{split}
\end{equation}
where $\nabla_{{q}} = \nabla_{{q}}f(q)$.

\begin{proof}
	Recall that the real scalar function $f(q)$ can also be treated as a real mapping $f(r_q)$, it follows that $\exp(-\lambda f(r_q))$ is concave. Since ${2}/{\eta}\leq\lambda$, it is easy to vertify that the function $h(r_q) \triangleq \exp(-\frac{2}{\eta} f(r_q))$ is also concave. Then by the concavity of $h(r_q)$, we have
	\begin{equation}
	\nonumber
	h(r_{q_1})\leq h(r_{q_2})+(\nabla_{r_{q_2}} h)^T(r_{q_1}-r_{q_2}).
	\end{equation}
	From \cite{opInQ}, we know that
	\begin{equation}
	\nonumber
	\nabla_{r_q}^T\Delta r_q = \nabla_{\underline{q}}^H\Delta \underline{q}.
	\end{equation}
	It then follows that
	\begin{equation}
	\nonumber
	h(q_1)\leq h(q_2)+(\nabla_{\underline{q_2}} h)^H(\underline{q_1}-\underline{q_2}),
	\end{equation}
	where $h(q) \triangleq \exp(-\frac{2}{\eta} f(q))$ and the
	augmented gradient $\nabla_{\underline{q}} h \triangleq
	-\frac{2}{\eta} \exp(-\frac{2}{\eta} f(q))\nabla_{\underline{q}}$
	according to the GHR calculus\cite{ghr}. Plugging these into the
	above equation results in
	\begin{equation}
	\nonumber
	f(q_1) \ge f(q_2)-\frac{\eta}{2}\log[1-\frac{2}{\eta} \nabla_{\underline{q_2}} ^H(\underline{q_1}-\underline{q_2})].
	\end{equation}
	We denote $z=\frac{2}{\eta}
	\nabla_{\underline{q_2}}^H(\underline{q_1}-\underline{q_2})$. Note
	that $\left|z\right|=\left|\frac{2}{\eta} \nabla_{\underline{q_2}}
	^H(\underline{q_1}-\underline{q_2})\right| \leq \frac{2}{\eta} GD
	\leq \frac{1}{4}$ and that for $\left|z\right|\leq \frac{1}{4}$,
	$-\log(1-z)\ge z+\frac{1}{4}z^2$, we have
	\begin{equation}
	\nonumber
	\begin{split}
	f(q_1) &\ge
	f(q_2)+\nabla_{\underline{q_2^*}}^H (\underline{q_1}-\underline{q_2})+\frac{1}{2\eta}(\underline{q_1}-\underline{q_2})^H\nabla_{\underline{q_2^*}}
	\nabla_{\underline{q_2^*}}^H(\underline{q_1}-\underline{q_2}).
	\end{split}
	\end{equation}
	Thus, we complete the proof of Lemma 7.
\end{proof}

Lemma 7 introduces an approximation of quaternion Taylor series
expansion using only the augmented quaternion gradient, which
gives the way to replace the augmented quaternion Hessian with the
matrix $A_{qt}$ defined in (\ref{qons}).

Before giving the next lemma, we introduce some characteristics
about the quaternion linear algebra, which differs from its real
and complex counterparts in many aspects, due to the
non-commutability of quaternion multiplication. For a quaternion
square matrix $A\in\mathbf{H}^{n\times n}$, we use $\lambda_i(A)$
to denote the standard eigenvalue of $A$, $\trace(A)$ to denote
the trace of $A$, $\left|A\right|_q$ to denote the $q$-determinant
of $A$ based on complex matrix representations \cite{qlabook}. We
use $\re\{\cdot\}$ to denote the real part of a quaternion scalar.
Then, for quaternion matrices $A,B \in \mathbf{H}^{n\times n}$, we
have \cite{qlapaper,qlabook}
\begin{gather}
\left|A\right|_q=\prod_{i=1}^{n}\left|\lambda_i\right|^2 \tag{13.a}, \\
\left|AB\right|_q=\left|BA\right|_q \tag{13.b}, \\
\left|A^{-1}\right|_q=\left|A\right|_q^{-1} \tag{13.c}, \\
\re\{\trace(AB)\}=\re\{\trace(BA)\}, \tag{13.d}
\end{gather}
and further, if $A$ are Hermitian matrix, i.e., $A^H=A$, we have
$\lambda_i(A) \in \mathbf{R}$ and that \cite{qlapaper}
\begin{gather}
\trace(A)=\sum_{i=1}^{n}\lambda_i(A). \tag{13.e}
\end{gather}

Based on (13), we have the following lemma.

\emph{Lemma 8}: Assume for all $t$, function $f(q):
\mathcal{K}\to\mathbf{R}$ has the property that  $\forall
q\in\mathcal{K},\|\nabla_{\underline{q}}f\|\leq G$. Then for the
definition of $A_{qt}=\sum_{i=1}^{t}\nabla_i\nabla_i^H+\varepsilon
I_{4n}$, we have
\begin{equation} \nonumber
\sum_{t=1}^{T}\nabla_t^H A_{qt}^{-1} \nabla_t = 
O(n\log T),
\end{equation}
where $\nabla_t=\nabla_{q_t}f_t$.

\begin{proof}
	From the augmentation property, we know that $\nabla_t^H A_{qt}^{-1}\nabla_t$ is a real scalar. According to (13.d) and (13.b), it then follows that
	\begin{equation} \nonumber
	\begin{split}
	\nabla_t^H A_{qt}^{-1}\nabla_t
	&= \re\{\trace(\nabla_t^H A_{qt}^{-1}\nabla_t)\} \\&
	= \re\{\trace(A_{qt}^{-1}\nabla_t\nabla_t^H)\} \\&
	= \re\{\trace(A_{qt}^{-1}(A_{qt}-A_{q(t-1)}))\} \\&
	=\re\{\trace(A_{qt}^{-\frac{1}{2}}(A_{qt}-A_{q(t-1)})A_{qt}^{-\frac{1}{2}})\}.
	\end{split}
	\end{equation}
	According to the definition of $A_{qt}$, it is easy to verify that
	the matrix
	$A_{qt}^{-\frac{1}{2}}(A_{qt}-A_{q(t-1)})A_{qt}^{-\frac{1}{2}}$ is
	Hermitian and positive definite. Thus its elements on the main
	diagonal are real scalars, and its standard eigenvalues are real
	positive scalars. By applying these facts, together with (13.e),
	we get
	\begin{equation} \nonumber
	\begin{split}
	\nabla_t^H A_{qt}^{-1}\nabla_t
	&= \trace(A_{qt}^{-\frac{1}{2}}(A_{qt}-A_{q(t-1)})A_t^{-\frac{1}{2}}) \\&
	= \trace(I-A_{qt}^{-\frac{1}{2}}A_{q(t-1)}A_{qt}^{-\frac{1}{2}}) \\&
	= \sum_{t=1}^{4n}[1-\lambda_t( A_{qt}^{-\frac{1}{2}}A_{q(t-1)}A_{qt}^{-\frac{1}{2}})] \\&
	\leq - \sum_{t=1}^{4n}\log[\lambda_t( A_{qt}^{-\frac{1}{2}}A_{q(t-1)}A_{qt}^{-\frac{1}{2}})] \\&
	= -\log[\prod_{t=1}^{4n}\lambda_t( A_{qt}^{-\frac{1}{2}}A_{q(t-1)}A_{qt}^{-\frac{1}{2}})].
	\end{split}
	\end{equation}
	Next by applying (13.a) and (13.c), we have that
	\begin{equation} \nonumber
	\begin{split}
	\nabla_t^H A_{qt}^{-1}\nabla_t
	\leq -\log \left|A_{qt}^{-\frac{1}{2}}(A_{q(t-1)})A_{qt}^{-\frac{1}{2}}\right|_q^\frac{1}{2} = \log[{\frac{\left|A_{qt}\right|_q}{\left|A_{q(t-1)}\right|_q}}]^{\frac{1}{2}}.
	\end{split}
	\end{equation}
	Summing the above for all $t$ yields
	\begin{equation} \nonumber
	\begin{split}
	\sum_{t=1}^{T}\nabla_t^H A_{qt}^{-1}\nabla_t
	&\leq \log[{\frac{\left|A_{qT}\right|_q}{\left|A_{q0}\right|_q}}]^{\frac{1}{2}}.
	\end{split}
	\end{equation}
	Finally, since $A_{qT}=\sum_{i=1}^{T}\nabla_i\nabla_i^H+\varepsilon I_{4n}$ and $\left\|\nabla_i\right\| \leq G$, the largest eigenvalue of $A_{qT}$ is at most $G^2T+\varepsilon$. Hence the $q$-determinant of $A_T$ can be bounded by $\left|A_{qT}\right|_q \leq (G^2T+\varepsilon)^{8n}$. Plugging this into the above inequality completes the proof.
\end{proof}

Based on Lemma 7 \& 8, we now have the following Lemma 9
for  Algorithm 2.

{\emph{Lemma 9}}: For the quaternion data sequence
$\{x_t\}_{t=1}^T$ generated by any qARMA model satisfying
Assumption 1-3, Algorithm 2 with loss function satisfying
Assumption 4, 5, 7, learning rate satisfying $\frac{1}{\eta} =
\frac{1}{2}\min\{\frac{1}{4GD},\lambda\}$, and initial matrix
$A_{q0}=\frac{\eta^2}{D^2} I_{4(p+m)}$ can generate an online
quaternion sequence $\{\gamma_t\}_{t=1}^T$ such that
\begin{equation}
\nonumber
\sum_{t=1}^{T}l_t^m(\gamma_t)-\min_{\gamma} \sum_{t=1}^{T}l_t^m(\gamma) = O((GD+\frac{1}{\lambda})(p+m)\log T).
\end{equation}
\begin{proof}
	Let $\gamma^\star=\arg\min_{\gamma} \sum_{t=1}^{T}l_t^m(\gamma)$
	denote the best qAR coefficient in hindsight, and denote
	$\nabla_{\gamma} = \nabla_{\gamma} l_t^m$. According to Lemma 7,
	for $\frac{1}{\eta}=\frac{1}{2}\min\{\frac{1}{4GD},\lambda\}$, we
	have
	\begin{equation}
	\begin{aligned}
	l_t^m(\gamma_t) - l_t^m(\gamma^\star)  &\leq \nabla_{\underline{\gamma_t^*}}^H(\underline{\gamma_t}-\underline{\gamma^\star})-\frac{1}{2\eta}(\underline{\gamma^\star}-\underline{\gamma_t})^H\nabla_{\underline{\gamma_t^*}}\nabla_{\underline{\gamma_t^*}}^H(\underline{\gamma^\star}-\underline{\gamma_t}).
	\end{aligned} \tag{14}
	\end{equation}
	Then according to the descent step, we have
	\begin{equation}
	\begin{split}\nonumber
	&(\underline{\phi_{t+1}}-\underline{\gamma^\star})^H A_{qt}(\underline{\phi_{t+1}}-\underline{\gamma^\star})=(\underline{\gamma_t}-\underline{\gamma^\star})^H A_{qt}(\underline{\gamma_t}-\underline{\gamma^\star})-{2}{\eta}\nabla_{\underline{\gamma_t^*}}^H(\underline{\gamma_{t}}
	-\underline{\gamma^\star})+{\eta^2}\nabla_{\underline{\gamma_t^*}}^H A_{qt}^{-1} \nabla_{\underline{\gamma_t^*}},
	\end{split}
	\end{equation}
	and according to the projection step, we have
	\begin{equation}
	\nonumber
	(\underline{\phi_{t+1}}-\underline{\gamma^\star})^H A_{qt}(\underline{\phi_{t+1}}-\underline{\gamma^\star}) \ge
	(\underline{\gamma_{t+1}}-\underline{\gamma^\star})^H A_{qt}(\underline{\gamma^{t+1}}-\underline{\gamma^\star}).
	\end{equation}
	Combining the above two steps yields
	\begin{equation}
	\nonumber
	(\underline{\gamma_{t+1}}-\underline{\gamma^\star})^H A_{qt}(\underline{\gamma_{t+1}}-\underline{\gamma^\star})\leq(\underline{\gamma_t}-\underline{\gamma^\star})^H A_{qt}(\underline{\gamma_t}-\underline{\gamma^\star}) -{2}{\eta} \nabla_{\underline{\gamma_t^*}} ^H(\underline{\gamma^{t}}
	-\underline{\gamma^\star})+{\eta^2} \nabla_{\underline{\gamma_t^*}}^H A_{qt}^{-1} \nabla_{\underline{\gamma_t^*}},
	\end{equation}
	and hence
	\begin{equation} \nonumber
	2\nabla_{\underline{\gamma_t^*}}^H(\underline{\gamma_t}-\underline{\gamma^\star})
	\leq {\eta}\nabla_{\underline{\gamma_t^*}}^H A_{qt}^{-1} \nabla_{\underline{\gamma_t^*}} +
	\frac{1}{\eta}(\underline{\gamma_t}-\underline{\gamma^\star})^H A_{qt}(\underline{\gamma_t}-\underline{\gamma^\star}) - \frac{1}{\eta}(\underline{\gamma_{t+1}}-\underline{\gamma^\star})^H A_{qt}(\underline{\gamma_{t+1}}-\underline{\gamma^\star}).
	\end{equation}
	By summing both sides of the above inequality for all $t$ and making some manipulation, we obtain that
	\begin{equation} \nonumber
	\begin{split}
	&2\sum_{t=1}^{T}\nabla_{\underline{\gamma_t^*}}^H(\underline{\gamma_t}-\underline{\gamma^\star})\\
	&\leq {\eta}\sum_{t=1}^{T}\nabla_{\underline{\gamma_t^*}}^H A_{qt}^{-1} \nabla_{\underline{\gamma_t^*}} + \frac{1}{\eta}\sum_{t=1}^{T}(\underline{\gamma_t}-\underline{\gamma^\star})^H \nabla_{\underline{\gamma_t^*}}\nabla_{\underline{\gamma_t^*}}^H(\underline{\gamma_t}-\underline{\gamma^\star})+ \frac{1}{\eta}(\underline{\gamma_1}-\underline{\gamma^\star})^H (A_1-\nabla_{\underline{\gamma_1^*}}\nabla_{\underline{\gamma_1^*}}^H)(\underline{\gamma_1}-\underline{\gamma^\star}).
	\end{split}
	\end{equation}
	Then by using the fact that $\|\underline{\gamma_1}-\underline{\gamma^\star}\|\leq D$ and that $A_1-\nabla_{\underline{\gamma_1^*}}\nabla_{\underline{\gamma_1^*}}^H = \frac{\eta^2}{D^2} I_{4(p+m)}$, we get that
	\begin{equation} \nonumber
	\begin{split}
	2\sum_{t=1}^{T}\nabla_{\underline{\gamma_t^*}}^H(\underline{\gamma_t}-\underline{\gamma^\star}) - \frac{1}{\eta}\sum_{t=1}^{T}(\underline{\gamma_t}-\underline{\gamma^\star})^H \nabla_{\underline{\gamma_t^*}}\nabla_{\underline{\gamma_t^*}}^H(\underline{\gamma_t}-\underline{\gamma^\star})\leq {\eta}\sum_{t=1}^{T}\nabla_{\underline{\gamma_t^*}}^H A_{qt}^{-1} \nabla_{\underline{\gamma_t^*}} + \eta
	\end{split}
	\end{equation}
	Combining it with (14) yields
	\begin{equation}
	\nonumber
	\begin{split}
	\sum_{t=1}^{T}(l_t^m(\gamma_t) - l_t^m(\gamma^\star)) \leq \frac{\eta}{2}\sum_{t=1}^{T}\nabla_{\underline{\gamma_t^*}}^H A_{qt}^{-1} \nabla_{\underline{\gamma_t^*}} + \frac{\eta}{2}.
	\end{split}
	\end{equation}
	According to Lemma 8, it follows that
	\begin{equation}
	\nonumber
	\begin{split}
	\sum_{t=1}^{T}(l_t^m(\gamma_t) - l_t^m(\gamma^\star)) = O(\frac{\eta}{2}(p+m)\log T).
	\end{split}
	\end{equation}
	Finally, since $\frac{1}{\eta} =
	\frac{1}{2}\min\{\frac{1}{4GD},\lambda\}$, we have
	$\eta\leq8(GD+\frac{1}{\lambda})$. Plugging this into the above
	equation gives the stated regret bound. Thus, we complete the
	proof of Lemma 9.
\end{proof}

Combining the results in Lemma 5 \& 9, we get the following Theorem 2, which
states a logarithmic bound on the regret of Algorithm 2.

{\emph{Theorem 2}}: For the quaternion data sequence
$\{x_t\}_{t=1}^T$ generated by any qARMA model satisfying
Assumption 1-3, Algorithm 2 with loss function satisfying
Assumption 4, 5, 7, learning rate satisfying $\frac{1}{\eta} =
\frac{1}{2}\min\{\frac{1}{4GD},\lambda\}$, and initial matrix
$A_{q0}=\frac{\eta^2}{D^2} I_{4(p+m)}$ can generate an online
quaternion sequence $\{\gamma_t\}_{t=1}^T$ such that
\begin{equation}
\nonumber
\sum_{t=1}^{T}l_t^m(\gamma_t)-\mathop{\min}_{\alpha,\beta} \sum_{t=1}^{T} \mbox{E}\left[f_t(\alpha,\beta)\right] = O((GD+\frac{1}{\lambda})(p+m)\log T).
\end{equation}

\section{Simulation Examples} \label{sec:5}
In this section, we present numerical examples to illustrate the
effectiveness of the proposed algorithms in both Guassian and
non-Guassian noise situations. We choose the squared loss as the
loss function, and use the MSE averaged over the past iterations
to evaluate the performance. For comparison, we also simulate the
multiple univariate ARMA-OGD and ARMA-ONS \cite{oarma} applied
component-wise, and their multichannel analogues ARMA-MOGD and
ARMA-MONS \cite{multichannel}. We average the results over 20
independent runs.

\emph{Example 1} (Guassian noise): In this example, we generate
the quaternion time series through a qARMA($4,2$) model where
$\alpha=[1.79-0.1i-0.2j,-1.85+0.1j-0.2k,1.27+0.2i+0.1k,-0.41-0.1i+0.1j]$
and $\beta=[0.9-0.2i+0.1j+0.3k,-0.5+0.5i-0.2k]$. The noise is
normally distributed as $\mathcal{N}(0,0.3^2)$, then for the
generated quaternion signals, the best possible qARMA predictor in
hindsight will suffer an average error rate of $4*0.3^2=0.36$.
According to the Algorithms, we choose $m=6$, i.e., $p+m=10$. The simulation result is shown in Fig. 1. As we see, both qARMA-ONS and
qARMA-OGD approach the optimum, while all the non-quaternion
algorithms fail to give satisfying predictions, which illustrates
the effectiveness and advantage of the proposed algorithms.

\begin{figure}[htbp]
	\centering
	\includegraphics[scale=0.5]{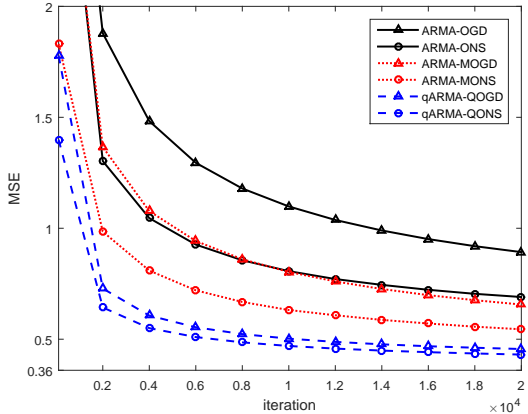}
	\caption{\label{fig:1} Simulation result of Example 1.}
\end{figure}

\emph{Example 2} (non-Guassian noise): In this example, we set the
condition to be the same as in Example 1, except that the noise is
distributed uniformly on $[-0.5,0.5]$. Then for the generated
quaternion signals, the best possible qARMA predictor in hindsight
will suffer an average error rate of 0.33. The simulation result
is shown in Fig. 2. As we see, again the quaternion algorithms
approach the optimal and outperform the non-quaternion algorithms.

\begin{figure}[htbp]
	\centering
	\includegraphics[scale=0.5]{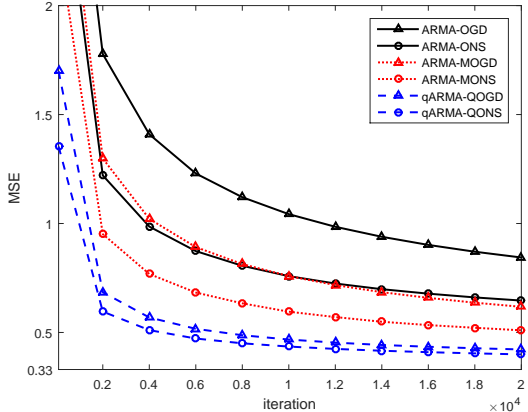}
	\caption{\label{fig:1} Simulation result of Example 2.}
\end{figure}

\section{Conclusion} \label{sec:6}
This paper proposed two online learning algorithms for the qARMA
model. We transformed the learning problem of qARMA into a full
information optimization task without explicit noise terms, and
then extended online gradient descent and Newtons methods to the
quaternion domain to solve the optimization problem. We further
gave theoretical analyses and simulation examples to show the
validity of this approach.

\end{document}